\documentclass[11pt]{article}

\usepackage[final]{acl}

\usepackage{times}
\usepackage{latexsym}
\usepackage{booktabs}
\usepackage{multirow}

\usepackage[T1]{fontenc}
\usepackage[table]{xcolor}
\usepackage{siunitx}
\usepackage[utf8]{inputenc}

\usepackage{adjustbox}

\usepackage{microtype}

\usepackage{inconsolata}

\usepackage{graphicx}
\usepackage{csquotes}

\usepackage{array}
\usepackage{collcell}
\usepackage{pgf}
\usepackage{etoolbox}

\usepackage{amsmath} 

\usepackage{tikz}
\usepackage{xcolor}
\usepackage{pifont}
\usepackage{arydshln}

\newcommand{\cmark}{\ding{51}}%
\newcommand{\xmark}{\ding{55}}%

\newcolumntype{L}{>{\small}l}

\definecolor{anchorcolor}{RGB}{255, 165, 0}   
\definecolor{acceptcolor}{RGB}{144, 238, 144}  
\definecolor{unstablecolor}{RGB}{255, 182, 193} 
\definecolor{graytoken}{RGB}{220, 220, 220}    
\definecolor{lcpaccept}{RGB}{100, 149, 237}

\newcommand{\tok}[2][white]{\colorbox{#1}{\strut\texttt{#2}}}

\newtoggle{cometpercent}
\toggletrue{cometpercent}   

\newtoggle{chrfpercent}
\toggletrue{chrfpercent}   

\newtoggle{seconds}
\toggletrue{seconds}         

\newtoggle{normalpercent}
\toggletrue{normalpercent}         

\newcommand{\cometfmt}[1]{%
  \if\relax\detokenize{#1}\relax %
  \else
    \iftoggle{cometpercent}{%
      \pgfmathparse{#1*100}%
      \pgfmathprintnumber[fixed, precision=2, zerofill]{\pgfmathresult}%
    }{#1}%
  \fi
}

\newcommand{\chrffmt}[1]{%
  \ifx#1-\relax
    -%
  \else
    \iftoggle{chrfpercent}{%
      \pgfmathparse{#1}%
      \pgfmathprintnumber[fixed, precision=2, zerofill]{\pgfmathresult}%
    }{#1}%
  \fi
}

\newcommand{\normalfmt}[1]{%
  \ifx#1-\relax
    -%
  \else
    \iftoggle{normalpercent}{%
      \pgfmathparse{#1*1}%
      \pgfmathprintnumber[fixed, precision=2, zerofill]{\pgfmathresult}%
    }{#1}%
  \fi
}

\newcommand{\latfmt}[1]{%
  \if\relax\detokenize{#1}\relax --%
  \else
    \iftoggle{seconds}{%
      \pgfmathparse{#1/1000}%
      \pgfmathprintnumber[fixed, precision=2, zerofill]{\pgfmathresult}%
    }{#1}%
  \fi
}

\newcommand{\latfmtfix}[1]{%
  \if\relax\detokenize{#1}\relax %
  \else
    \iftoggle{seconds}{%
      \pgfmathparse{#1/1000}%
      \pgfmathprintnumber[fixed, precision=2, zerofill]{\pgfmathresult}%
    }{#1}%
  \fi
}

\newcolumntype{Q}{>{\collectcell\cometfmt}c<{\endcollectcell}}
\newcolumntype{N}{>{\collectcell\normalfmt}c<{\endcollectcell}}
\newcolumntype{W}{>{\collectcell\latfmt}c<{\endcollectcell}}
\newcolumntype{Z}{>{\collectcell\latfmtfix}c<{\endcollectcell}}
\newcolumntype{C}{>{\collectcell\chrffmt}c<{\endcollectcell}}

\usepackage{xcolor}

\definecolor{mutedgreen}{HTML}{1B5E20} 
\definecolor{mutedred}{HTML}{B71C1C}   

\usepackage{xparse} 

\NewDocumentCommand{\splitlatdelta}{ >{\SplitArgument{1}{_}}m }{%
  \processlatdelta#1%
}

\NewDocumentCommand{\processlatdelta}{ m m }{%
  \if\relax\detokenize{#1}\relax 
  \else
    \iftoggle{seconds}{%
      \pgfmathparse{#1/1000}%
      \pgfmathprintnumber[fixed, precision=2, zerofill]{\pgfmathresult}%
    }{%
      \pgfmathparse{#1}%
      \pgfmathprintnumber[fixed, precision=2, zerofill]{\pgfmathresult}%
    }%
    \IfNoValueF{#2}{%
      \pgfmathparse{#2 >= 0 ? 1 : 0}%
      \ifnum\pgfmathresult=1
        \def\deltacolor{mutedgreen}%
      \else
        \def\deltacolor{mutedred}%
      \fi
      $_{\textcolor{\deltacolor}{%
        \iftoggle{seconds}{%
          \pgfmathparse{#2/1000}%
          \pgfmathprintnumber[fixed, precision=2, zerofill, showpos]{\pgfmathresult}%
        }{%
          \pgfmathparse{#2}%
          \pgfmathprintnumber[fixed, precision=2, zerofill, showpos]{\pgfmathresult}%
        }%
      }}$%
    }%
  \fi
}

\newcommand{\latdeltafmt}[1]{\splitlatdelta{#1}}

\newcolumntype{D}{>{\collectcell\latdeltafmt}c<{\endcollectcell}}

\title{MLLP-VRAIN UPV system for the IWSLT 2026\\Simultaneous Speech Translation task}
\newcommand{\upvdir}{Machine Learning and Language Processing, VRAIN, Universitat Politècnica de València}
\author{
   Jorge Iranzo-Sánchez*, Gerard Mas-Mollà* \\
   {\bf Adrià Giménez, Jorge Civera, Albert Sanchis, Alfons Juan}
  \\
  \upvdir\\
  \texttt{\{jorirsan,gemamol\}@upv.es}
}

\begin{document}
\maketitle
\begin{abstract}
This work describes the participation of the MLLP-VRAIN research group in the shared task of the IWSLT 2026 Simultaneous Speech Translation track. Our submission utilizes the recently released Parakeet and Qwen 3.5 models to create a robust, cascaded solution for long-form SimulST through the use of adaptive \enquote{black-box} policies. We explore relaxations of these policies to achieve better quality-latency trade-offs.
Compared to last year, we participate on all language directions. In addition to this, for the En$\rightarrow${De, It, Zh} directions we also participate in this year's new context track employing a combination of ASR word-boosting and a RAG mechanism of offline pre-translated exemplars to guide generation and enrich our system with domain-specific context. Finally, we provide a detailed latency analysis of our system.
Compared to last year, results on the MCIF En$\rightarrow$De test set shows a substantial quality improvement of +5.82 XCOMET-XL. Our context track processing further improves performance by +1.03.
\end{abstract}

\section{Introduction}
\begin{figure}[t]
    \centering
    \includegraphics[width=\linewidth]{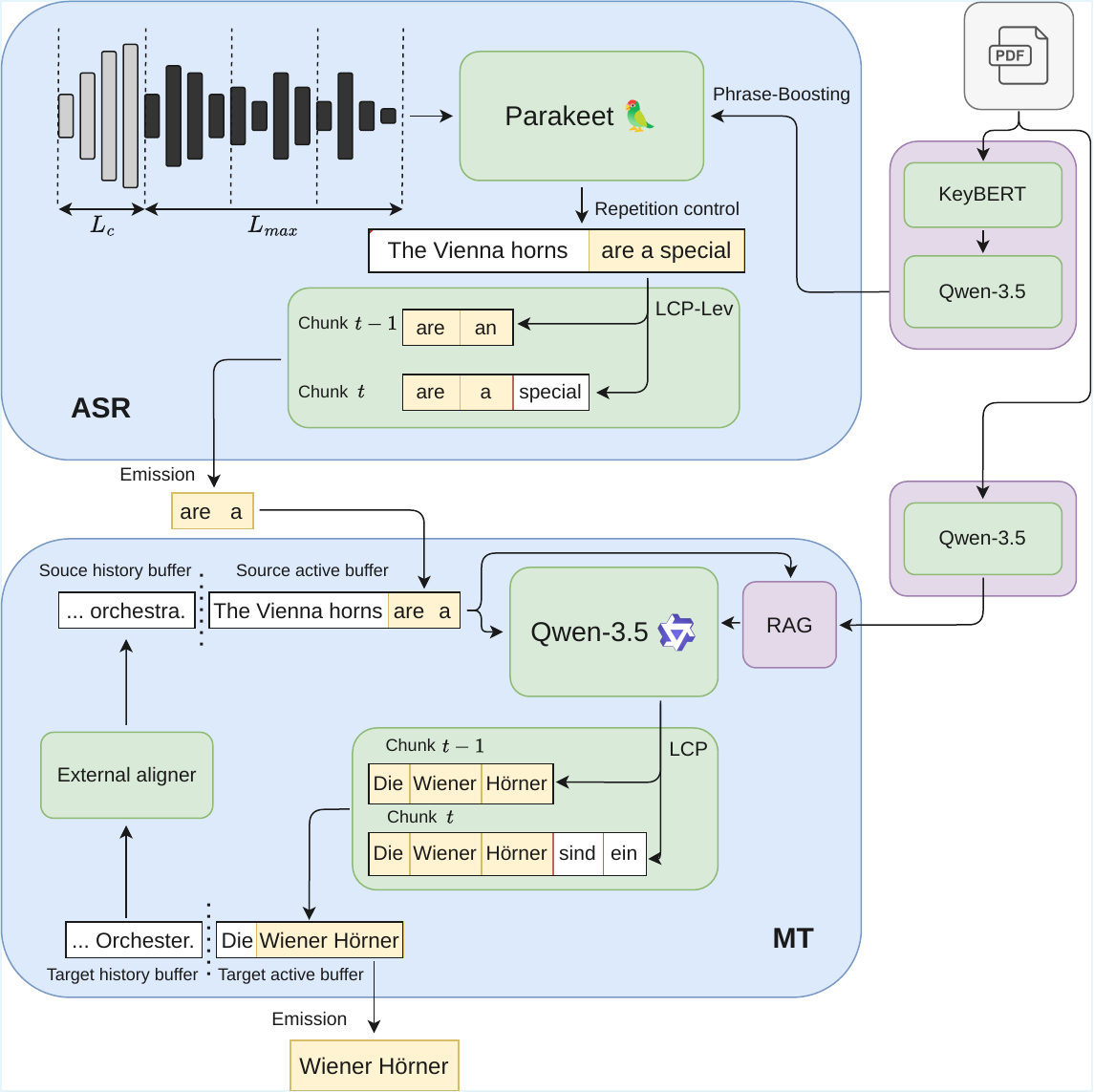}
    \caption{System diagram of our cascaded system for the SimulST track}
    \label{fig:system_overview}
\end{figure}

In this paper we describe the participation of
the MLLP-VRAIN research group in the shared
tasks of the 23th International Conference on Spoken Language Translation (IWSLT)~\cite{proceedings-2026-iwslt}.
Building on our previous participation on last year IWSLT SimulST Track~\cite{iranzo-sanchez-etal-2025-mllp}, we focus on cascaded solutions for SimulST. This choice is motivated by the IWSLT results from last year~\citep{agostinelli-etal-2025-findings}, where cascaded systems achieved the best performance, as well as the strong results reported on the recently introduced Hearing2Translate~\citep{papi2025hearingtranslateeffectivenessspeech} benchmark and the flexibility that cascaded approaches give us in the choice of our components. 
We also believe that the creation of a strong cascaded system may show which audio and text components are optimal for the creation of a derived SpeechLLM further down the line. 
Figure \ref{fig:system_overview}~shows the overall architecture of our system which will be described in more detail in the following sections.

This year, we participate in all language directions and latency regimes. Furthermore, we participate in the newly introduced extra context track for the En$\rightarrow${De, It, Zh} directions, proposing mechanisms to leverage this context for both the ASR and MT components. 
The evaluation metrics we use are XCOMET-XL~\cite{GuerreiroRSCCM24} and chrF~\cite{popovic-2015-chrf,machacek-etal-2023-mt} for translation quality
and LongYAAL~\cite{polak2026betterlatenevermetaevaluation} for latency.

\section{Surface Level Black Box Policies for SimulST}
Black-box emission policies are a well-established approach in SimulST. They do not rely on direct access to internal model information and can therefore be applied to any offline model without additional training, covering both fixed policies such as Wait-k~\cite{ma-etal-2019-stacl} and Hold-n~\cite{LiuSN20} and adaptive ones such as Longest Common Prefix (LCP)~\cite{liu20s_interspeech}. Notably, 
\cite{gerard_streamingasr} recently showed that the combination of offline SOTA  ASR models with black box policies achieved competitive results on streaming ASR benchmarks compared to more specialized solutions. Recent winners of past editions of IWSLT have also demonstrated the effectiveness of LCP for the creation of SOTA SimulST systems~\cite{polak-etal-2022-cuni,polak-etal-2023-towards,machacek-polak-2025-simultaneous}.

One of the most widely used adaptive policies is the previously mentioned LCP, which accepts the longest common prefix across consecutive model generations as a valid output. However, in practice, LCP often causes high latency spikes when the system has a high degree of oscillation between tokens across generations. However, in many cases, these oscillations 
have little impact on the final quality. Last year, we relaxed our ASR LCP policy to account for this phenomenon by accepting tokens based on a Levenshtein distance threshold, a method we refer to as LACP~\cite{iranzo-sanchez-etal-2025-mllp}.
We propose that this relaxation can be taken further to achieve lower latency at a modest quality trade-off. We refer to this further relaxation as \textit{Soft LCP} (SLCP). SLCP is motivated by two observations. First, the Ratcliff/Obershelp (RO) pattern recognition algorithm~\cite{ratcliff1988pattern} could provide a more suitable string similarity measure than Levenshtein distance for this task. Second, there frequently exist \enquote{anchor} tokens that remain stable across generations even when surrounding tokens vary slightly without affecting the final quality. Figure~\ref{fig:slcp-anchor} illustrates in more detail the SLCP policy. The idea is to identify \enquote{anchor} tokens via RO and greedily accept all preceding tokens, propagating committed output more frequently than regular LCP would allow. We define $\gamma$
as the maximum allowable gap (in tokens) between unstable tokens for anchor propagation, and $\sigma$ as the minimum similarity score for a token to qualify as an anchor. For all experiments in this work, we set $\gamma=3$ and $\sigma=0.6$.

\newcommand{\ctok}[2][graytoken]{%
  {\setlength{\fboxsep}{1.5pt}\colorbox{#1}{\texttt{#2}}}%
}

\begin{figure*}[ht]
\centering
\renewcommand{\arraystretch}{1.5}
\resizebox{0.8\textwidth}{!}{%
\begin{tabular}{c l}
\toprule
\textbf{Gen.} & \textbf{Hypothesis Tokens} \\
\midrule
$g_1$ &
  \tok[graytoken]{the} \;
  \tok[graytoken]{ether} \;
  \tok[graytoken]{near} \;
  \tok[graytoken]{Plasencia} \\
$g_2$ &
  \tok[lcpaccept]{the} \;
  \tok[graytoken]{weather} \;
  \tok[graytoken]{in} \;
  \tok[anchorcolor]{Palencia} \;
  \tok[unstablecolor]{reminds} \;
  \tok[unstablecolor]{me} \;
  \tok[unstablecolor]{of} \;
  \tok[anchorcolor]{Valencia} \;
  \tok[graytoken]{and} \\
\midrule
\textbf{Out} &
  \tok[acceptcolor]{the} \;
  \tok[acceptcolor]{weather} \;
  \tok[acceptcolor]{in} \;
  \tok[acceptcolor]{Palencia} \;
  $\longleftarrow$ \textit{accepted via anchor propagation} \\
\bottomrule
\end{tabular}%
}
\caption{SLCP anchor propagation example with maximum gap $\gamma=2$ and $\sigma=0.6$.
\ctok[anchorcolor]{Palencia} and \ctok[anchorcolor]{Valencia} are identified as possible \textbf{anchor tokens}
of \ctok[graytoken]{Plasencia} with scores 0.82 and 0.70 respectively.
\ctok[anchorcolor]{Valencia} is not a valid anchor, since the length of the
chunk \ctok[unstablecolor]{reminds me of} is $>\gamma$, remaining unstable and not being committed.
All preceding tokens of \ctok[anchorcolor]{Palencia} are greedily accepted, emitting \ctok[acceptcolor]{the weather in Palencia}.
If using LCP, only the first token \ctok[lcpaccept]{the} would have been accepted. Note that \ctok[graytoken]{weather} is also a valid anchor with respect \ctok[graytoken]{ether}.
}
\label{fig:slcp-anchor}
\end{figure*}

\section{ASR Component}

\paragraph{Speech Foundational Models} Our ASR component was chosen based on the results of public ASR systems on the benchmarks available at the HuggingFace Open ASR Leaderboard~\cite{openasr}. Based on the language pairs considered in the competition and attending to our computing limitations, we needed a lightweight multilingual model capable of producing high quality transcriptions under streaming conditions. We finally selected Parakeet~\cite{sekoyan2025canary1bv2parakeettdt06bv3efficient}
\footnote{Model: \href{https://huggingface.co/nvidia/parakeet-tdt-0.6b-v3}{nvidia/parakeet-tdt-0.6b-v3}}
as our ASR component. Our decision of using Parakeet can be explained by two main reasons. The first reason is that, as a multilingual model, it supports Czech ASR, and so it allows us to participate in the Cs$\rightarrow$En direction. The second reason is that it is a lightweight model with only 0.6B parameters, and as such it allows the usage of heavier LLMs as MT systems. Apart from that, since Parakeet already achieves competitive results compared to other ASR systems on the development data provided by the organizers, we did not perform any finetuning process to the model.

The adaptation of Parakeet to streaming was performed following~\cite{gerard_streamingasr}, where three main components are applied to perform online decoding. First, incremental data ingestion is managed by an acoustic buffer that receives fixed-length chunks of size $L_c$. By defining a maximum buffer size $L_{max}$, the input audio buffer behaves as a sliding window, growing cumulatively until $L_{max}$ is reached. From this point on, whenever a new chunk is added, it pushes the oldest chunk out of the buffer. Following this idea, the acoustic input $X_t$ at any decoding step $t$ can be formally expressed as $X_t = [\text{max}(0,t\cdot L_C - L_{max}), t\cdot L_c]$. Then, since the entire acoustic buffer is fed to the model at every decoding step, and since we set that $L_c \ll L_{max}$, the model is forced to process acoustic information that has already been processed in previous decoding steps, thus leading to repetitions in the output transcription. This phenomenon is mitigated by a timestamp-based repetition control. We leverage Parakeet's capacity to predict token durations to keep track of emission times at the output of the model, which allows us to filter any repetition based on the information of previous decoding steps. Finally, once the output has been filtered, it is added to an output buffer governed by an emission policy. Additionally, we used the ALSD++ beam search decoding implementation of NeMo~\cite{GrigoryanBAXLG25} with beam size 32.

\begin{figure}
    \centering
    \includegraphics[width=\linewidth]{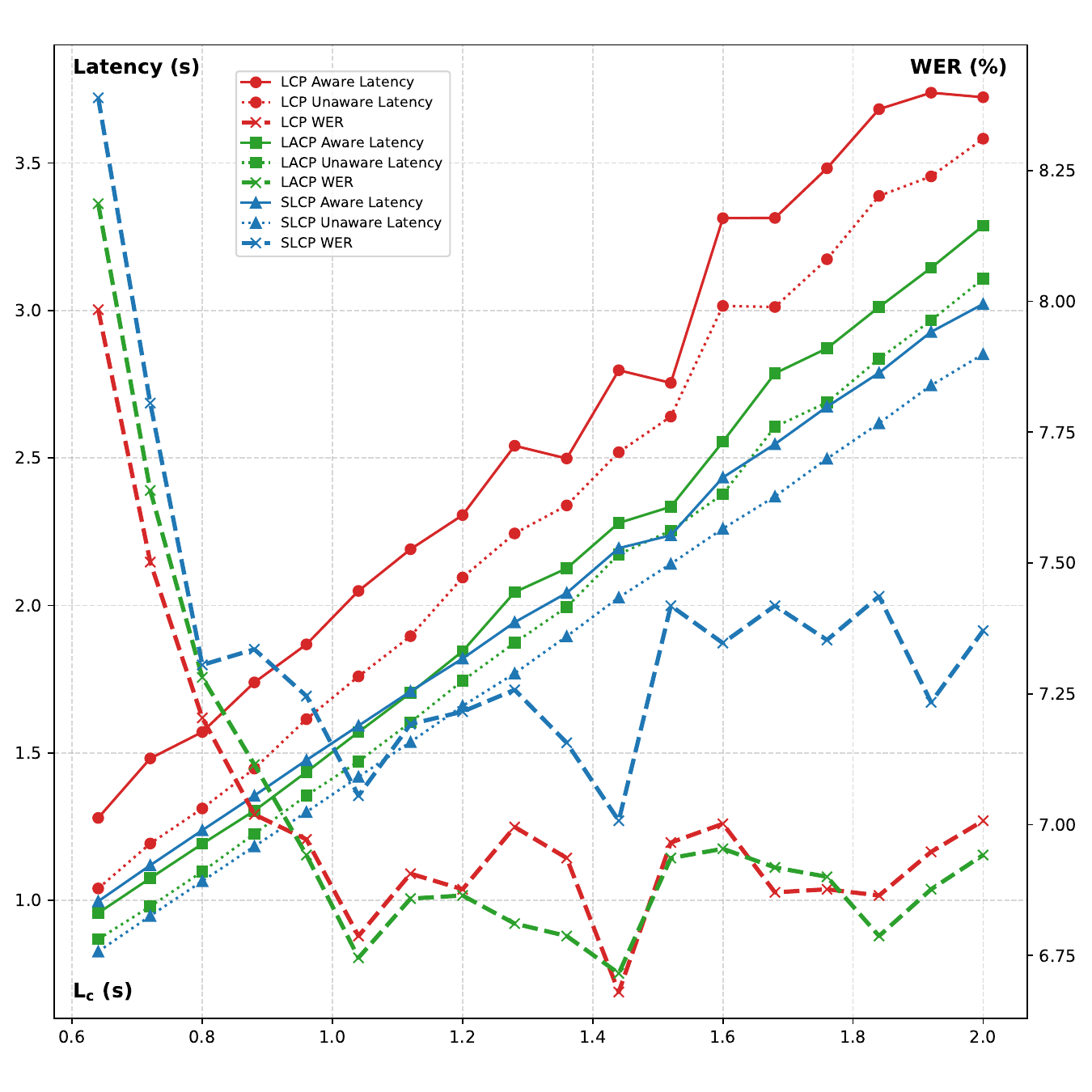}
    \caption{Latency (left) vs. WER (right) trade-off of $L_c = \{\text{0.64},\dots,\text{2.00}\}$ sweep in MCIF.}
    \label{fig:asr_policy}
\end{figure}

\paragraph{Streaming ASR with emission policies} The candidate policies considered for the ASR model were LCP, LACP and SLCP. Fixed policies such as Wait-$k$ and Hold-$n$ were discarded based on the results of preliminary informal experiments, as they showed poorer WER/latency trade-offs, particularly in lower latency configurations. Regarding LACP, we set the Levenshtein threshold to $\tau=2$ following~\cite{gerard_streamingasr} and the findings of our participation last year. The candidate policies were tested by sweeping the chunk size $L_c$ from 0.64 to 2.00 seconds and measuring both computational aware and unaware latency values. The latency figures are computed by aligning the output transcripts with an external HMM-based system using the TLK toolkit~\cite{tlk}. Then, we compute the latency values using these alignments and the system emission timestamps. All experiments were performed on a node with a NVIDIA RTX 4090 GPU and an Intel Core 10920X CPU. As seen in the results plotted in Figure~\ref{fig:asr_policy}, LCP and LACP stand out in terms of WER, consistently yielding better results than those of SLCP. As for the latency results, LACP and SLCP perform similarly on low latency configurations, with SLCP yielding the best results in high latency configurations. In light of the results, we selected LACP as the best policy, as it achieves competitive WER results while maintaining low latency figures.

\section{MT Component} 
\paragraph{LLMs: newer, bigger, better} 
In the context of SimulST, we are restricted in our choice of foundational models, as we need to be able to run them in real time (RTF <1) to have a true streaming model. Last year we made use of an encoder-decoder approach by using NLLB~\cite{nllb}, since we found it to be a good middle-ground in terms of model size, speed and performance. However, recent WMT evaluation have shown the performance of encoder-decoders such as NLLB to be subpar compared to state of the art LLMs for offline MT~\cite{kocmi-etal-2024-findings,kocmi-etal-2025-findings}. Consequently, we were motivated this year to explore more in depth current LLMs for SimulST, following the plethora of recent work that demonstrate their effectiveness on this task~\cite{koshkin-etal-2024-llms,koshkin-etal-2024-transllama,raffel-etal-2024-simultaneous,agent_simt,seed_liveinterp}. 
 
We conducted an initial survey of publicly available open-weight LLMs and selected several candidates for a preliminary, informal evaluation: HuanYan-MT-1.5~\cite{hy-mt1.5}, EuroLLM~\cite{ramos2026eurollm22btechnicalreport}  Tower+~\cite{rei2025towerplus},  TranslateGemma~\cite{finkelstein2026translategemma} and Qwen 3.5~\cite{qwen3.5}.
Preliminary experiments revealed stability issues in several models of the first model families (e.g., frequent refusals and oscillatory outputs), leading us to focus on TranslateGemma and Qwen 3.5
for further exploration as our primary machine translation models. Table~\ref{tab:qwen_gemma} presents preliminary XCOMET, chrF and YAAL results for the 4B variants of both used model families, as well as for Qwen3.5 9B and 27B. For the latter two, different quantization methods are also tested to be able to use these model sizes on the <24GB consumer grade GPUs we have available. Results show comparable performance across the two families. This led us to select the Qwen 3.5 family as the backbone of our MT component. This decision was further reinforced upon discovering that TranslateGemma had been trained on a fixed prompt template and thus exhibited poor robustness to prompt variations and external context insertion. 
Based on the results, we will use the quantized 27B for our final system submission. We also leveraged the 9B-fp8 variant on part of the experimentation.

\begin{table}[t]
\centering
\small
\begin{adjustbox}{max width=\columnwidth}
\begin{tabular}{lcccc}
\toprule
\textbf{Model} & \multicolumn{2}{c}{\textbf{YAAL} $\downarrow$} & \multirow{2}{*}{\textbf{XCOMET} $\uparrow$} & \multirow{2}{*}{\textbf{chrF} $\uparrow$} \\
\cmidrule(lr){2-3}
 & \textbf{CU} & \textbf{CA} & & \\
\midrule
\quad UPV IWSLT25                    & 3.18 & 3.43 & 86.85 & 60.23 \\
\midrule
\quad TranslateGemma (4B)            & 3.04 & 3.20 & 89.45 & 57.92 \\
\cmidrule(lr){1-5}
\quad Qwen 3.5 (4B)                  & 2.99 & 3.33 & 89.48 & 58.06 \\
\quad Qwen 3.5 (9B)                  & 2.94 & 3.39 & 89.57 & 58.55 \\
\quad Qwen 3.5 (9B, fp8)             & 2.92 & \textbf{3.19} & 90.19 & 57.79 \\
\quad Qwen 3.5 (27B, fp8)            & 2.84 & 4.21 & 90.86 & \textbf{59.25} \\
\quad Qwen 3.5 (27B, int4\footnotemark)           & \textbf{2.87} & 3.46 & \textbf{91.09} & 58.89 \\
\bottomrule
\end{tabular}
\end{adjustbox}
\caption{MCIF En$\rightarrow$De quality--latency  results of initial long-form streaming systems.  LLMs based models use an emission policy of Hold-3 and for the MT and the ASR component of our last year submission.}
\label{tab:qwen_gemma}
\end{table}

\footnotetext{Model: \href{https://huggingface.co/Intel/Qwen3.5-2B-int4-AutoRound}{Intel/Qwen3.5-2B-int4-AutoRound}}

\paragraph{MT Buffer control} Last year, we finetuned our MT component to emit \enquote{sentinel} tokens following~\citet{iranzo-sanchez-etal-2024-segmentation} which served to indicate a target side end of sentence. This would then trigger a history buffer update that would identify the corresponding source-size end of sentence by obtaining an alignment through the usage of cross-attention maps as proxy alignments~\cite{li-etal-2019-word}. This year, we get rid of this mechanism.
This decision is based on two observations. First, in our previous system, where the source and output streams were cased (unlike the original paper, which assumed lowercase), the system typically generated a sentinel token after a strong punctuation mark (\texttt{!?.}) was emitted. Thus, if we can identify when this punctuation is generated, we can directly use it to trigger the alignment mechanism. Second, while we use LLMs, we still require source-target alignments. To our knowledge, obtaining reliable alignments from text-based LLMs using attention maps remains an open question. Although some works suggest that alignments can be achieved through a discrete or generative approach by self-prompting the model~\cite{mao-yu-2024-tuning}, we decided to use a lightweight external aligner, similar to how external CTC alignments are used in ASR to obtain timestamps. This avoids potential model hallucinations during the alignment task.
In our experimentation we run SimAlign~\cite{jalili-sabet-etal-2020-simalign} with XLM-Roberta Base ($\sim$125M) as a backend model~\cite{conneau-etal-2020-unsupervised} quantized to int8 and running on CPU to minimize GPU memory usage and computing overhead. We set the maximum history buffer to 20 sentences or 1024 words (or characters for En$\rightarrow$Zh) and eject the oldest sentence when this limit is surpassed in either the source or target buffer.

\paragraph{Decoding } Due to model size and to keep computational costs at a reasonable range, we decided to use greedy search for our MT component as the decoding algorithm. We also explored the use of Minimum Bayes Risk (MBR) decoding for both ASR and MT component; however, mixed results led us to discard this approach in our final submissions. The results of this exploration are detailed in Appendix ~\ref{sec:appendix_mbr}.

\paragraph{Prevention of Catastrophic Failure}
We identify two rare cases in which the MT system fails catastrophically and cannot recover. In certain configurations, early termination may occur within a document: the system stops emitting tokens for the current segment due to an overconfident prediction that produces strong punctuation. This, in turn, causes the EOS token to dominate subsequent emissions, even if the source stream continues to grow.
To mitigate this issue, we allow the system to rewrite the last two previously emitted tokens when such a condition is detected. With this mechanism in place, we no longer observe this phenomenon, and the introduced flickering remains minimal.
In addition, we observe occasional oscillatory hallucinations. To address these, we adopt the temperature fallback mechanism proposed in Whisper \cite{RadfordKXBMS23}. Unlike the original work, we only trigger the temperature fallback if the gzip compression ratio of the emitted tokens exceeds 2.4. 
\section{Cascaded system}

\begin{figure}
    \centering
    \includegraphics[width=\linewidth]{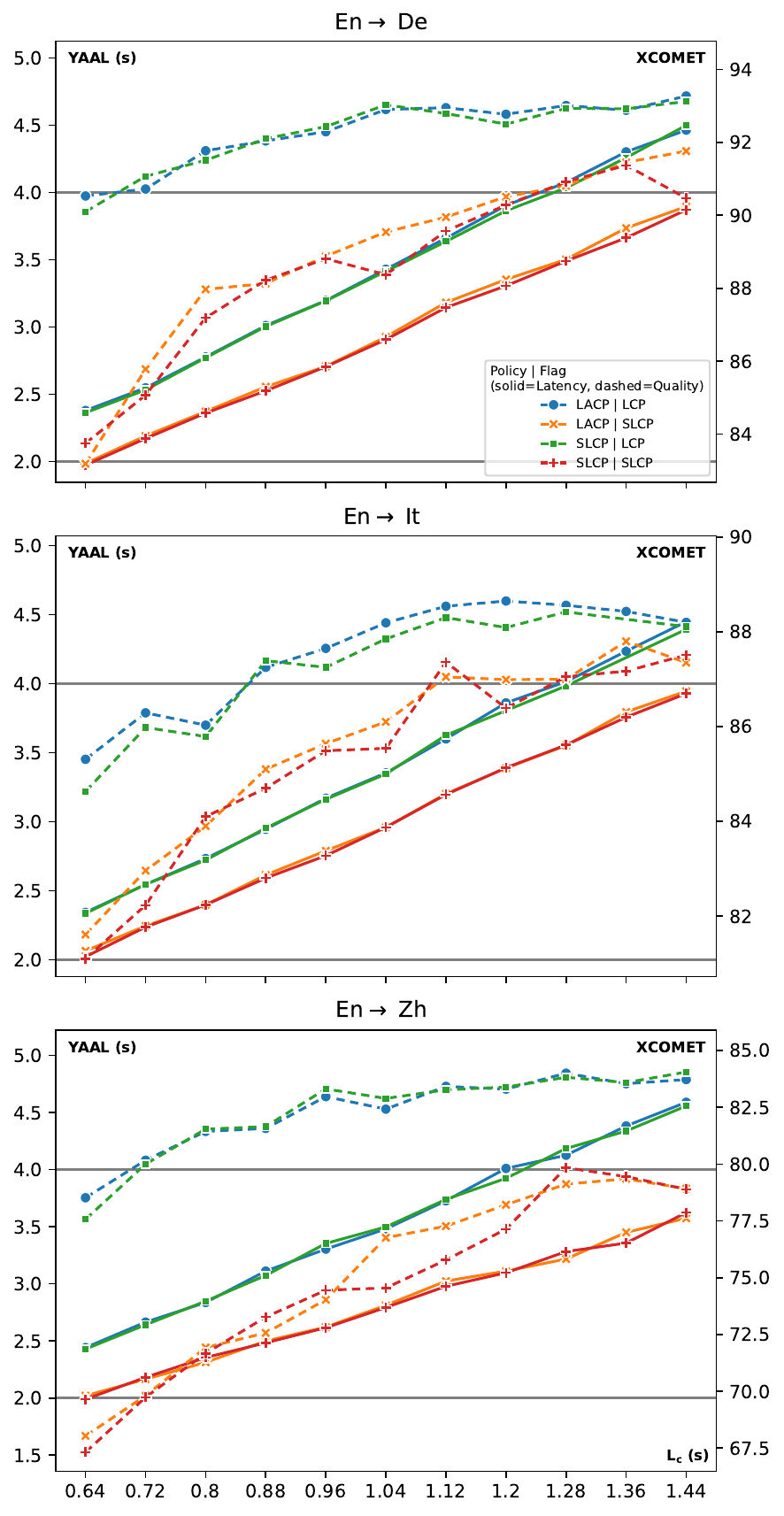}
    \caption{YAAL (left) vs. XCOMET (right) trade-off of $L_c \in \{\text{0.64},...,\text{1.44}\}$ sweep on the MCIF~\cite{papi2026mcif} IWSLT 2026 test set  with Qwen 27B.
    }
    \label{fig:multi_sweep}
\end{figure}

\paragraph{Optimal buffer sizes} Figure~\ref{fig:multi_sweep} presents results for all language directions of MCIF, comparing the usage of LACP/SLCP in the ASR component and LCP/SLCP in MT.
Focusing first on the ASR component, we observe that LACP and SLCP yield very similar latency and quality across different values of $L_c$, with LACP showing a slight overall advantage. 
For the MT component, comparing LCP and SLCP, we find that SLCP can achieve a considerable reduction in average YAAL, with latency improvements of approximately 0.3--1 seconds compared to LCP. However, these gains come at the cost of noticeable drops in XCOMET, particularly at smaller $L_c$ values. 
In last year's evaluation, our system achieved significantly lower latency than the winning system, but at the expense of translation quality. Human evaluation, however, showed a clear preference for the higher-latency system. This suggests that in our case, mid-range chunk size configurations with LCP may be preferable to lower-latency SLCP alternatives by human preference.
Based on these observations, we ultimately adopt LACP for our ASR systems and LCP for our MT system, leaving further investigation of SLCP for more latency-constrained scenarios. 
Regarding the acoustic chunk size, we select $L_c=$ 1.04s for our high latency configurations in all language directions, as the quality seems to peak around this value.

\begin{figure}
    \centering
    \includegraphics[width=\linewidth]{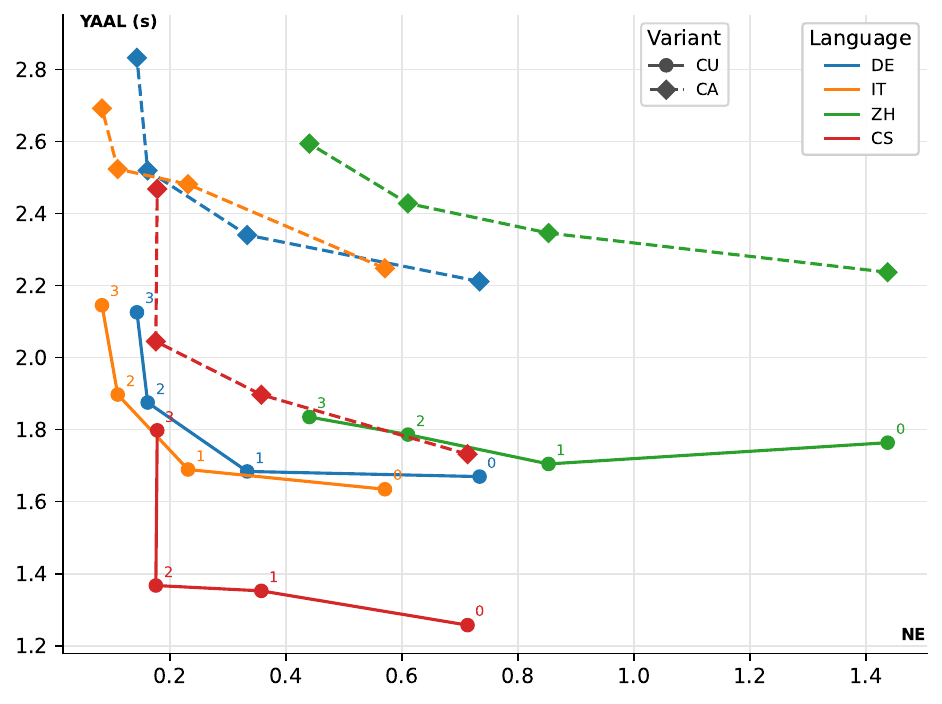}
    \caption{Normalized erasure (x-axis) vs. YAAL (y-axis) trade-off for Mask-$k$ with $k \in \{0, \dots, 3\}$ on the LCP speculative emission using Qwen3.5 9B with $L_c=0.64$s}
    \label{fig:holdn}
\end{figure}

\paragraph{Re-translation for Low Latency} As shown in Figure~\ref{fig:multi_sweep}, for all our tested configurations there is no models for which the YAAL <2 seconds restrictions for the low latency track is valid. 
As such, we adopt a simple mask-$k$ re-translation based approach~\citep{ArivazhaganCTMB20,arivazhagan-etal-2020-translation} on the MT component. By applying this technique, we are able to reduce YAAL of systems with low $L_c$ values to participate in this latency regime.
More specifically, at each step, we take the non committed suffix resulting from the LCP based policy, remove the last $k$ tokens and consider the resulting output suffix as a \enquote{speculative} emission which is not committed to the internal translation buffer, and for which tokens can be overwritten in the next generation. Figure~\ref{fig:holdn} shows both computational aware and unaware YAAL-Normalized Erasure trade-off across all language directions for Qwen 3.5 variant, $L_c=$ 0.64 and $k \in \{0, \dots, 3\}$. 
Based on the results, we select $k=2$ as the optimal choice, as it obtains the best generalizable latency-flickering ratio across all languages directions and fix $L_c=$ 0.64 for our low latency systems.~\footnote{\citet{arivazhagan-etal-2020-translation} considers a model to have \enquote{few revisions} if NE < 0.2.} 

\section{Context Track}
We also participate in this year extra context track which allows participants to use additional information from associated paper PDFs.

\subsection{Word boosting for ASR} 
For ASR, we leverage the efficient GPU-accelerated Phrase-Boosting (GPU-PB)~\citep{gpupb_2025_1143461} implementation for the Parakeet models backed by Nvidia NeMo to guide ASR generation by shallow fusion from an extracted keyword list from the given PDFs. 
For keyword list extraction, we make use of a two-step strategy. First, we use KeyBERT~\citep{grootendorst2020keybert}\footnote{Model: \href{https://huggingface.co/sentence-transformers/all-MiniLM-L6-v2}{sentence-transformers/all-MiniLM-L6-v2}} to get an initial set of keywords. Then, we reuse the same Qwen 3.5 model used as our MT backbone to refine the keywords extracted in the first step.
Also, compared to the organizer's baseline, we make use of the whole document instead of just the title, author's list and abstract section. 
More specifically, in our extraction pipeline, all paper sections except the references are first extracted and cleaned with some formatting regexes. Then, the full cleaned text is chunked into overlapping segments. These segments are then fed to KeyBERT, which extracts the initial list of keywords. Finally, the keywords are passed to the LLM to be refined. 
Additionally, we tested two levels of granularity for the application of GPU-PB: dataset and document-level.

\begin{figure}
    \centering
    \includegraphics[width=\linewidth]{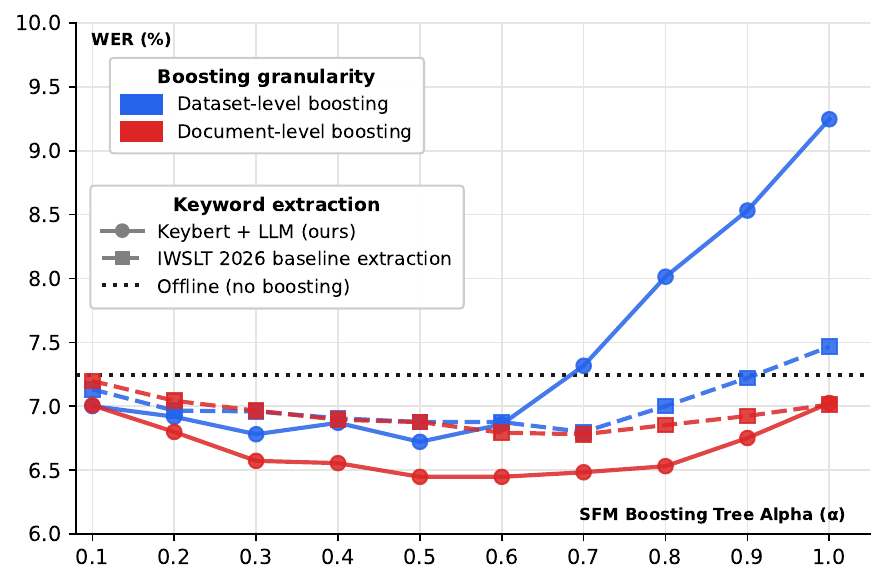}
    \caption{WER (x-axis) vs.\ SFM boosting tree alpha ($\alpha$) (y-axis) on the MCIF IWSLT 2026 test set for greedy search and $L_c=0.96s$.
    }
    \label{fig:wer_vs_alpha}
\end{figure}

Figure~\ref{fig:wer_vs_alpha} shows results sweeping across GPU-PB $\alpha$ interpolation parameter, comparing our keyword extraction method versus the organizers baseline at different levels of granularity.
We see that word-boosting at the document level obtains lower WER results overall compared to the dataset level. Furthermore, we consider $\alpha=$ 0.6 to be optimal on the MCIF dev set, as it reduces WER results from 7.2 to 6.4. We set this $\alpha$ value for our final model.

\subsection{RAG with lexical retrieval for MT}
Since the context track PDFs only provide source-language information, we provide the MT model with additional contextual guidance by pretranslating 
the document at the sentence level, creating an offline translation memory that can be queried at runtime. 
This component is intended to provide look-ahead hints about upcoming content,
while helping to preserve consistent and accurate terminology. Our hypothesis is that this approach can improve both translation quality and latency, particularly in scenarios where the system lacks full context and may otherwise struggle to disambiguate terms correctly.
For our retrieval mechanism, we take the list of source and target sentence translation pairs. Then, before starting decoding, we generate a BM25s~\cite{bm25s} index per document with default parameters by creating a lowercased and lex-normalized copy of the source sentences.
Then at run time, at each timestep we query the index with the current source sentence content and retrieve the indices of the top-$k$ best matches $r_k$. We then use $r_k$ to retrieve the best translation pairs and inject them as context into our prompt. Our selection for a lexical based approach is based on the demonstrated effectiveness of using BM25 for domain adaptation in offline translation by~\citet{agrawal-etal-2023-context}. 
In addition to this, the low cost of BM25s allows us to run the query on the CPU and avoid the training and the higher inference cost compared to a neural based solution such as that of RAAST~\cite{luo2026rasstfastcrossmodalretrievalaugmented}.
We tried three different configurations to inject the retrieved sentences: at the header position before the system prompt, after the source sentence context and before the source sentence context. From these three configurations, we make use of the latter, as we observed that with the other two, the model had a tendency to start hallucinating additional source and target pairs. 

Table~\ref{tab:rag_k_merged} reports YAAL and XCOMET results obtained by sweeping over different top-$k$ values for MCIF En$\rightarrow${De, It, Zh} with $L_c=$ 0.96, comparing them to base context-less systems and ASR word-boosted ones. As it can be observed, incorporating the RAG mechanism consistently improves XCOMET scores while maintaining latency comparable to both context-free systems and ASR word-boosted baselines. Regarding the number of retrieved exemplars $k$, the relationship between performance gains and quality varies across systems. Overall, we find in other reduced sweeps of $L_c$ with $k \in \{2, 5\}$ that increasing $k$ beyond two does not tend to lead to further improvements and can even plateau or slightly degrade performance by starting to retrieve irrelevant exemplars for the current active sentence. Based on these findings, we set $r_k=$ 2 for all final systems in the context track.

\begin{table}[h]
\centering
\begin{adjustbox}{max width=\columnwidth}
\begin{tabular}{cc QQQ ZZZ}
\hline
\multirow{2}{*}{\textbf{WB}} &
\multirow{2}{*}{\textbf{RAG-$k$}} &
\multicolumn{3}{c}{\textbf{XCOMET} $\uparrow$} &
\multicolumn{3}{c}{\textbf{YAAL \iftoggle{seconds}{(s)}{(ms)}} $\downarrow$} \\
\cmidrule(lr){3-5} \cmidrule(lr){6-8}
& &
\multicolumn{1}{c}{\textbf{De}} &
\multicolumn{1}{c}{\textbf{It}} &
\multicolumn{1}{c}{\textbf{Zh}} &
\multicolumn{1}{c}{\textbf{De}} &
\multicolumn{1}{c}{\textbf{It}} &
\multicolumn{1}{c}{\textbf{Zh}} \\
\hline
\xmark & \xmark
  & 0.9242 & 0.8777 & 0.7969	
  & 3401.2888	 & 3340.2817 & 3468.8132 \\
\cmark & \xmark
  & 0.9269 & 0.8802    & 0.8140 
  & 3493.2370 & 3367.2954	 & 3578.5628  \\
\midrule
\cmark & 1
  & 0.9299 & 0.8838 & 0.8169
  & 3445.6897 & 3322.4517 & 3521.4737 \\
\cmark & 2
  & 0.9301 & 0.8819 & 0.8220
  & 3414.9740 & 3319.0971 & 3550.8815 \\
\cmark & 3
  & 0.9294 & 0.8870 & 0.8167
  & 3516.8830 & 3404.6127 & 3468.9439 \\
\cmark & 4
  & 0.9306 & 0.8850 & 0.8227
  & 3361.1388 & 3388.3810 & 3655.2410 \\
\cmark & 5
  & 0.9328 & 0.8896 & 0.8294
  & 3521.5311 & 3367.1701 & 3661.0937 \\
\hline
\end{tabular}
\end{adjustbox}
\caption{Sweep for MT RAG system across $k$ for MCIF set with $L_c=0.96$ and Qwen 3.5 9B.
WB denotes ASR Word Boost; RAG-$k$ indicates retrieved exemplars.}
\label{tab:rag_k_merged}
\end{table}

\section{On SimulST Latency Scores}
\paragraph{True latency, \enquote{macro} average latencies and oracle offsets}  To ensure that our system latencies would have a similar performance in real use cases and reflect user-perceived latency (UPL), we calculate latency scores of our complete pipeline by calculating alignments of our final configurations. For the MT component, we make use of latency metric based on the definition of \citet{polak2026betterlatenevermetaevaluation} by using forced alignment of the audio and source references and then aligning to the translation hypothesis\footnote{CTC based aligners from WhisperX~\cite{BainHHZ23} and SimAlign \cite{jalili-sabet-etal-2020-simalign} }. We refer to this metric as TrueLatency.

\begin{table*}[h]
\centering
\begin{adjustbox}{max width=0.95\textwidth}

\begin{tabular}{lcQZDDDc}
\hline
\multirow{2}{*}{\textbf{Latency}} & 
\multirow{2}{*}{\textbf{Context}} & 
\multicolumn{1}{c}{\multirow{2}{*}{\textbf{XCOMET} $\uparrow$}} & 
\multicolumn{1}{c}{\multirow{2}{*}{$\mathbf{YAAL_{Macro}}$ $\downarrow$}} &
\multicolumn{3}{c}{$\mathbf{YAAL_{Micro+EndOffset}}$ $\downarrow$} & 
\multirow{2}{*}{\textbf{NE} $\downarrow$} \\
\cmidrule(lr){5-7}
& & & & 
\multicolumn{1}{c}{\textbf{Mean}} & 
\multicolumn{1}{c}{\textbf{P50}} & 
\multicolumn{1}{c}{\textbf{P99}} & \\
\hline
\multicolumn{8}{c}{\textbf{En-De}} \\
\hline
\multirow{2}{1cm}{LOW}  & \xmark  & 0.9026 & 1889.87 & 1514.09_23.52 & 1393.52_108.48 & 4186.43_1364.67 & 0.17 \\
                        & \cmark  & 0.9205 & 1894.45 & 1526.22_95.51 & 1422.64_143.36 & 4200.06_1498.48 & 0.21 \\
\hdashline
\multirow{2}{1cm}{HIGH} & \xmark  & 0.9267 & 3410.89 & 2989.88_147.87 & 2902.71_276.29 & 6304.69_854.95 & 0.00 \\
                        & \cmark  & 0.9370 & 3414.46 & 3000.22_177.73 & 2889.79_314.21 & 6570.47_661.52 & 0.00 \\
\hline
\rowcolor{gray!12}\multicolumn{8}{c}{\textbf{En-It}} \\
\hline
\rowcolor{gray!12}                        & \xmark  & 0.8516 & 1887.21 & 1535.51_13.32 & 1449.91_79.59 & 4018.54_1370.11 & 0.15 \\
\rowcolor{gray!12}\multirow{-2}{1cm}{LOW} & \cmark  & 0.8703 & 1896.80 & 1542.35_61.86 & 1443.22_99.78 & 4372.74_1143.86 & 0.18 \\
\hdashline
\rowcolor{gray!12}                         & \xmark  & 0.8797 & 3356.09 & 2959.91_74.75 & 2866.00_189.50 & 6234.76_956.43 & 0.00 \\
\rowcolor{gray!12}\multirow{-2}{1cm}{HIGH} & \cmark  & 0.8936 & 3417.68 & 2999.33_129.06 & 2912.33_198.67 & 6285.93_902.70 & 0.00 \\
\hline
\multicolumn{8}{c}{\textbf{En-Zh}} \\
\hline
\multirow{2}{1cm}{LOW}  & \xmark  & 0.7846 & 1796.87 & 1613.32_-204.42 & 1477.72_-130.72 & 4776.64_1544.76 & 0.38 \\
                        & \cmark  & 0.8212 & 1820.76 & 1650.35_-192.69 & 1489.35_-87.85 & 5170.87_1287.56 & 0.53 \\
\hdashline
\multirow{2}{1cm}{HIGH} & \xmark  & 0.8279 & 3444.68 & 3253.14_-218.58 & 3148.71_-69.71 & 6765.19_1280.81 & 0.00 \\
                        & \cmark  & 0.8456 & 3548.02 & 3360.05_-263.65 & 3229.06_-105.06 & 7466.02_895.28 & 0.01 \\
\hline
\rowcolor{gray!12}\multicolumn{8}{c}{\textbf{Cs-En}} \\
\hline
\rowcolor{gray!12}LOW  & \xmark  & 0.7760 & 1565.00 & 1071.89_441.55 & 1110.81_341.19 & 5336.97_3697.93 & 0.18 \\
\hdashline
\rowcolor{gray!12}HIGH & \xmark  & 0.8277 & 2793.00 & 1991.50_614.72 & 2552.40_353.60 & 7263.61_3278.13 & 0.00 \\
\hline
\end{tabular}
\end{adjustbox}
\caption{Final evaluation results across all language pairs. Subindices of $\mathbf{YAAL_{Micro+EndOffset}}$ submetrics indicate the corresponding $\Delta( \mathbf{TrueLatency_{Micro}} -\mathbf{YAAL_{Micro+EndOffset}})$.NE indicates Normalized Erasure. ~\label{tab:final_results}}
\end{table*}

\begin{figure}[h]
    \centering
    \includegraphics[width=\linewidth]{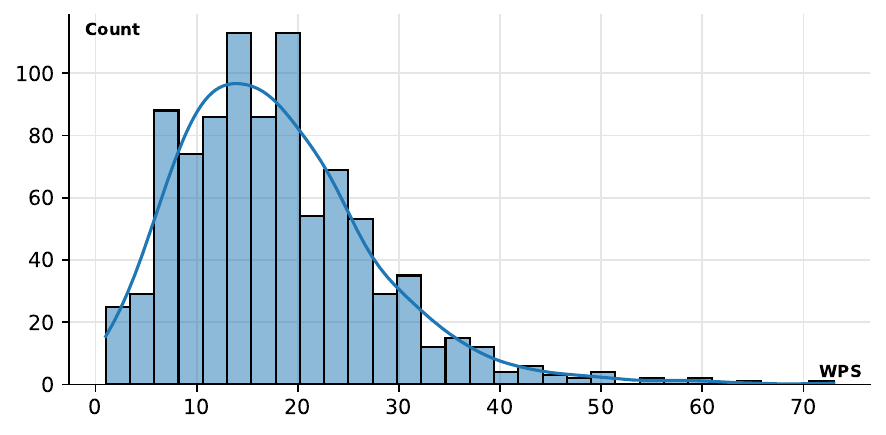}
    \caption{Words per sentence (WPS) histogram of MCIF En$\rightarrow$De target.}
    \label{fig:sentence_dis}
\end{figure}

\begin{figure}[h]
    \centering
    \includegraphics[width=\linewidth]{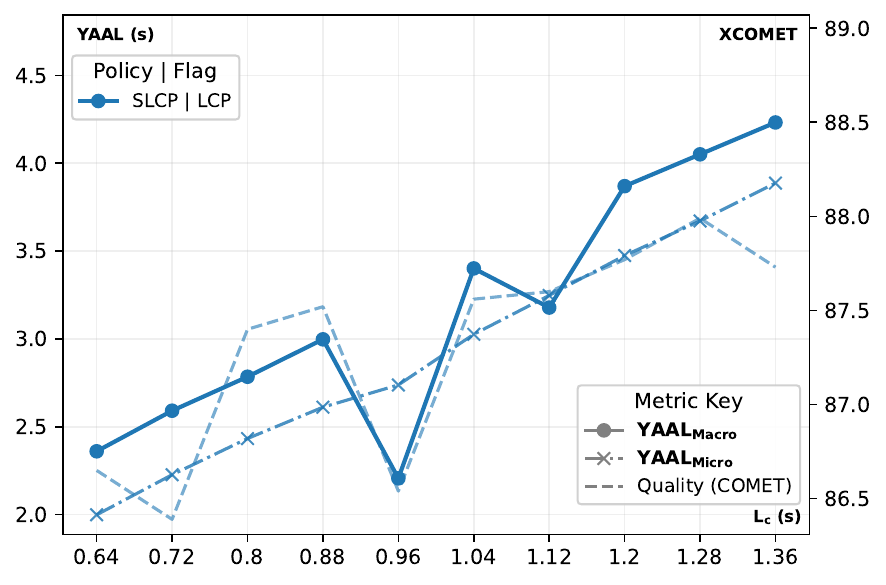}
    \caption{Example of early $L_c$ sweep with Qwen-9B SCLP+LCP for MCIF En$\rightarrow$It
    without whisper-like temperature fallback 
    where $L_c=0.96$ and $L_c=1.12$ present early \enquote{end of stream} failure cases. The $\mathbf{YAAL_{Macro}}$ gets artificially decreased due the negative latencies, while $\mathbf{YAAL_{Micro}}$ is more robust to this type of noise.}
    \label{fig:failure_case_micro_macro_comp}
\end{figure}

During this evaluation process, we identified three problems with current latency metrics. First, SimulST latency metrics are currently reported as a \textit{macro \textbf{average} of average token latencies} per sentence. We argue that in practice, this makes latency dependent on reference target segmentation and sentence length, distorting UPL. As an example, Figure~\ref{fig:sentence_dis} shows the source word length distribution of the MCIF dataset, where it can be seen that, by taking the \enquote{macro},
latency on longer sentences may be under-represented.
Our second identified problem lays on the way that AL based metrics calculate word delays with respect to the reference "wait-0" oracle.
When adapting text based AL metrics to source speech, the delay for a word is the difference between the emission time and the oracle assigned $(t-1/r)$, with $r$ being the length ratio between target and source sequences. This can also be interpreted as taking the \textbf{start} emission time of equally distributed source words. 
This mismatches standard ASR latency calculations, which measure the difference between hypothesis and reference end delays~\footnote{For example, see \href{https://github.com/MyrtleSoftware/caiman-asr/blob/0a69969c0a80a3bf786b4e05418d7f27f5cc780c/training/caiman_asr_train/latency/measure_latency.py}{Caiman ASR} and \href{https://github.com/ufal/asr_latency}{UFAL asr\_latency} script. It is worth noting that contrary to AL based metrics, ATD does take source end emission times on its formulation.}.
Our final identified problem is that 
current macro-level AL metrics are highly sensitive to alignment errors.
For instance, if a faulty system stops emitting prematurely,
the sentence aligner of YAAL may force-align single words from the last sentences to missing sentences, generating extreme negative delays that may distort the system's real latency.
We observe that taking the macro average in this cases greatly skews the YAAL scores, 
while the micro average smooths noisy negative delays, yielding a more realistic latency score for the functional part of the inference.
Figure~\ref{fig:failure_case_micro_macro_comp} shows an example of this phenomenon of faulty inferences in Qwen3.5 9B. Configurations with $L_c=$ 0.96 and $L_c=$ 1.12 show how YAAL calculated at the macro level in these cases artificially reduces latency with respect to the expected YAAL score that correlates with the increase of $L_c$, while YAAL at the micro level properly captures the expected linearity and behavior of the model.
In addition to all of this, this negative delay phenomenon can be easily overlooked, as common checks for empty sentence alignments will not report this cases.

\section{Final Results}
Following the previous section, for our final systems reported in Table~\ref{tab:final_results}, in addition to standard macro, start oracle emission YAAL, XCOMET and NE, we report the YAAL average at the micro level with oracle end offsets alongside the corresponding deltas with respect to our calculated TrueLatency. We also report median and p99 following the recommendations of~\cite{iranzo-sanchez-etal-2025-going} to ensure the robustness of our systems and give a better picture of latency distribution beyond the mean. 
For our final systems, $\mathbf{YAAL_{Macro}}$ latencies hover the $\sim$1.9 and $\sim$3.5 second mark for MCIF directions and 1.5 and 2.8 for Cs$\rightarrow$En for the low and high latency regimes. In terms of quality, compared to models configurations of this year organizer baselines on MCIF with similar YAAL scores~\footnote{Baseline models with 0.64s and 1.28s chunk size.}, we obtain substantial improvements, with $\Delta\textbf{XCOMET}^{Low}_{De,It,Zh}=(+13.5,+16.7,+3.0)$ and $\Delta\textbf{XCOMET}^{High}_{De,It,Zh}=(+7.6,+9.0,+2.3)$ for low and high latency respectively. Versus the improved context track baselines, we also maintain substantial gains of $\Delta\textbf{XCOMET}^{Low+Ctx}_{De,It,Zh}=(+14.4,+17.7,+7.2)$ and $\Delta\textbf{XCOMET}^{High+Ctx}_{De,It,Zh}=(+7.4,+9.7,+3.2)$.
We can also observe that our final models'
$\mathbf{YAAL_{Micro+EndOffset}}$ are very similar to the obtained $\mathbf{TrueLatency_{Micro}}$, alongside reasonable median and p99 values, which lead us to affirm that our final models are robust and their latency will probably reflect real observed UPL. We do note that for Cs
$\rightarrow$En, bigger $\Delta$ gaps appear compared to the MCIF language pairs.

\section{Limitations}
Several limitations of this work should be acknowledged. First, our exploration of relaxed LCP policies was limited due to time constraints. LACP was not evaluated as an MT emission policy, and the sensitivity analysis of SLCP parameters $\gamma$ and $\sigma$ were selected on previous small scale experiments. It is possible that per-language tuning of these hyperparameters could yield better latency--quality trade-offs for the remaining language directions, which we leave to future work.
Second, also due to time constraints, policy exploration in ASR was entirely conducted in English ASR. Since we transferred the best English configuration to Czech ASR, our hope is that our results for the Czech - English translation pair could be further improved with a language-specific policy exploration.
Third, our system is restricted to cascaded architectures. While this choice is empirically motivated by strong results on recent benchmarks, it may forgo potential gains from tighter integration of acoustic and linguistic information. We acknowledge that SpeechLLM-based approaches via a  modality adapter~\cite{verdini25_interspeech} represent a promising alternative, and our decision not to explore them here is primarily driven by the limited availability of in-domain training data for this track and the computational cost of bridging the modality gap in such architectures.
Finally, the computational budget available to us constrained several design choices. Our experiments were conducted mostly consumer-grade GPUs with at most 24GB of memory, which prevented us from evaluating larger unquantized models and from running MBR decoding at a scale that would be competitive with greedy decoding in terms of real-time factor. 

\section*{Acknowledgments}
We would like to specially thank the IWSLT Simultaneous Speech track organizers, especially Katsuhito Sudoh and Victor Agostinelli, for providing us with last years logs of the SimulST track. We also acknowledge the usage of large language model based tools to assist our writing and proofreading process of this paper.
The research leading to these results has received funding from EU4Health
Programme 2021--2027 as part of Europe's Beating Cancer Plan under Grant
Agreements nos. 101056995 and 101129375; and from the Government of
Spain's grant PID2021-122443OB-I00 funded by
MICIU/AEI/\allowbreak10.13039/\allowbreak5011\-00011033 and by
``ERDF/EU'', grant PDC2022-133049-I00 funded by
MICIU/AEI/\allowbreak10.13039/\allowbreak501100011033 and by the
``European Union NextGenerationEU/PRTR'',
and grant PRE2022-103662 funded by 
MICIU/AEI/\allowbreak10.13039/\allowbreak501100011033 and by “ESF+”.
The authors gratefully
acknowledge the financial support of Generalitat Valenciana under project
IDIFEDER/\allowbreak2021/\allowbreak059.


\appendix

\section{Minimum Bayes Risk Decoding Study}~\label{sec:appendix_mbr}

While Minimum Bayes Risk (MBR) decoding was popular during both the statistical ASR era~\cite{GoelB00} and early machine translation research~\cite{kumar-byrne-2004-minimum}, it has recently seen renewed interest in offline MT~\cite{eikema-aziz-2022-sampling}.

This resurgence is largely driven by the emergence of strong neural evaluation metrics, which can outperform standard beam search while avoiding known issues such as the beam search curse~\cite{murray-chiang-2018-correcting,yang-etal-2018-breaking}. To the best of our knowledge, this constitutes the first study exploring the use of MBR in the SimulST setting, as prior work has largely restricted MBR to offline ASR and MT scenarios~\cite{abs-2510-19471,wang-etal-2025-nyas,li-etal-2025-kits}.

We leverage the MBR implementations provided by the \texttt{mbrs} library~\cite{deguchi-etal-2024-mbrs} and experiment with multiple evaluation metrics. Due to computational constraints, we primarily focus on XCOMET-lite~\cite{larionov-etal-2024-xcomet} and chrF \footnote{\url{https://github.com/jvamvas/fastChrF}}. We also evaluated chrF++~\cite{popovic-2017-chrf}, default BLEU via sacreBLEU~\cite{post-2018-call}, and PartialCOMET~\cite{zouhar-etal-2026-early} in earlier experiments, but observed similar or worse performance at higher computational cost. For hypothesis generation, we use epsilon sampling~\cite{hewitt-etal-2022-truncation,freitag-etal-2023-epsilon} with $\epsilon=0.02$ and $\tau=1.0$. For all applicable metrics, we make use of Reference Aggregation to speed up MBR~\cite{denero-etal-2009-fast,vamvas-sennrich-2024-linear}

Table~\ref{tab:mbr_mt} reports results for configurations feasible on a single NVIDIA RTX 4090. For the Qwen models, chrF with $n=32$ performs comparably to greedy decoding, but at a significantly higher computational cost. This ultimately led us to discard MBR for our final submission.

The table also includes results for the IWSLT 2025 UPV system, where we replace the RALCP policy and force the system to always commit outputs. This setup highlights an interesting property of MBR when adapting offline methods to SimulST: it mitigates hallucinations and reduces the tendency of mode-seeking decoding algorithms to emit empty outputs. While the original submission utilized RALCP to address these issues, removing the emission policy causes both greedy and beam search decoding to produce divergent target content which end up in inference failures. In contrast, MBR naturally prevents this behavior, enabling stable decoding without requiring additional control policies.

We also explored applying offline MBR to the context track. However, generating large numbers of hypotheses ($k$) resulted in significantly slower decoding, with real-time factors exceeding 1 relative to dataset duration. As a result, we discarded this approach for the context track as well.

\begin{figure}[h]
\centering
\includegraphics[width=\linewidth]{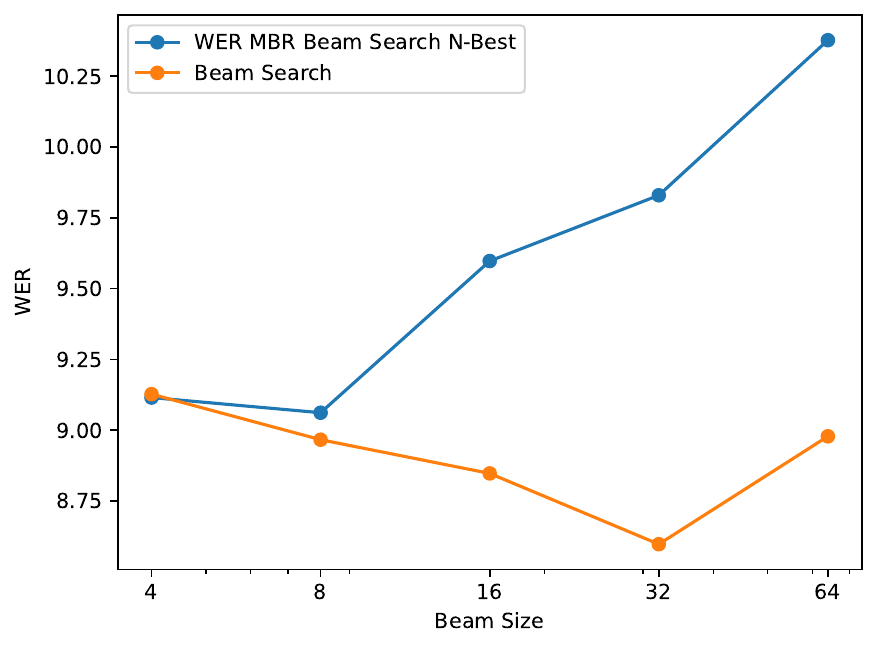}
\caption{WER of beam search vs.\ MBR re-ranking of $n$-best hypotheses for Parakeet with $L_c=0.96$ on MCIF across different values of $n$\label{fig:wer_mbr}}
\end{figure}

Finally, we evaluated a standard MBR re-ranking approach for ASR over $n$-best hypotheses generated via beam search. However, this method consistently yielded negative results, as shown in Figure~\ref{fig:wer_mbr}.
Overall, we did not adopt MBR in neither the ASR nor the MT component in our final system due to its computational cost and limited benefits, although it remains an interesting direction for future study.

\begin{table*}[p]
\centering
\small
\begin{tabular}{lclccccccc}
\toprule
\multirow{2}{*}{\textbf{Model}} & 
\multirow{2}{*}{\textbf{Policy}} & 
\multirow{2}{*}{\textbf{MBR Metric}} & 
\multirow{2}{*}{\textbf{Samples}} & 
\multicolumn{2}{c}{\textbf{YAAL (s)} $\downarrow$} & 
\multirow{2}{*}{\textbf{XCOMET} $\uparrow$} & 
\multirow{2}{*}{\textbf{chrF} $\uparrow$} & 
\multirow{2}{*}{\textbf{BLEU} $\uparrow$} \\
\cmidrule(lr){5-6}
& & & & \textbf{CU} & \textbf{CA} & & & \\
\midrule

\multirow{5}{*}{Qwen 3.5 (4B)} 
& \multirow{5}{*}{Hold-3} 
& Greedy & 1 & 2.99 & 3.33 & \textbf{89.48} & 58.06 & 24.75 \\
\cmidrule(lr){3-9}
& & \multirow{2}{*}{chrF} & 16 & 3.05 & 3.96 & 87.43 & 57.15 & 23.83 \\
& & & 32 & 3.10 & 4.52 & 87.15 & 58.01 & 24.66 \\
\cmidrule(lr){3-9}
& & XCOMET-lite & 8 & 3.06 & 3.96 & 85.33 & 51.40 & 17.51 \\

\midrule

\multirow{5}{*}{Qwen 3.5 (9B)} 
& \multirow{5}{*}{Hold-3} 
& Greedy & 1 & 2.94 & 3.39 & \textbf{89.57} & 58.55 & 25.58 \\
\cmidrule(lr){3-9}
& & \multirow{2}{*}{chrF} & 16 & 2.87 & 4.22 & 87.52 & 57.98 & 23.18 \\
& & & 32 & 2.89 & 4.87 & 89.55 & 59.22 & 24.66 \\
\cmidrule(lr){3-9}
& & XCOMET-lite & 8 & 2.82 & 4.04 & 88.51 & 53.14 & 17.38 \\

\midrule

\multirow{5}{*}{IWSLT 25 UPV} 
& \multirow{5}{*}{Write All}
& Greedy & \multicolumn{6}{c}{\xmark\;Unstable, results in an inference failure} \\
& & Beam Search & \multicolumn{6}{c}{\xmark\;Unstable, results in an inference failure} \\
\cmidrule(lr){3-9}
& & chrF & 64 & 1.96 & 2.43 & 76.46 & 56.19 & 20.98 \\
& & chrF++ & 64 & 1.99 & 5.34 & 76.74 & 54.90 & 20.36 \\
& & BLEU & 64 & 2.23 & 4.91 & 76.70 & 54.10 & 22.85 \\
& & PartialComet & 64 & 2.22 & 2.71 & 76.29 & 45.85 & 10.28 \\

\bottomrule
\end{tabular}
\caption{Comparison of Greedy vs. MBR decoding for Qwen 3.5 variants hold-3 variants and IWSLT 25 UPV without RALCP for MCIF En$\rightarrow$De.}
\label{tab:mbr_mt}
\end{table*}

\end{document}